\documentclass[11pt]{article}
\usepackage[hyperref]{acl2020}
\usepackage{times}
\usepackage{latexsym}
\usepackage{graphicx}

\usepackage[T1]{fontenc}
\usepackage[utf8]{inputenc}
\usepackage{devanagari} 
\usepackage{microtype}
\usepackage{multirow}
\aclfinalcopy 

\begin{document}
\title{BITS Pilani at HinglishEval: Quality Evaluation for Code-Mixed Hinglish Text Using Transformers}


\author{Shaz Furniturewala, Vijay Kumari, Amulya Ratna Dash, Hriday Kedia,\\ \textbf{Yashvardhan Sharma}\\
  BITS Pilani,Pilani,Rajasthan \\
  \texttt{(f20200025,p20190065)@pilani.bits-pilani.ac.in} \\
  \texttt{(p20200105,f20190964,yash)@pilani.bits-pilani.ac.in} 
 }

\maketitle
\begin{abstract}
Code-Mixed text data consists of sentences having words or phrases from more than one language. Most multi-lingual communities worldwide communicate using multiple languages, with English usually one of them. Hinglish is a Code-Mixed text composed of Hindi and English but written in Roman script. This paper aims to determine the factors influencing the quality of Code-Mixed text data generated by the system. For the HinglishEval task, the proposed model uses multi-lingual BERT to find the similarity between synthetically generated and human-generated sentences to predict the quality of synthetically generated Hinglish sentences.

\end{abstract}

\section{Introduction}

The term "Code-Mixing" refers to mixing words or phrases from different languages into a single text or speech utterance. It embeds linguistic units from one language, such as phrases, words, and morphemes, into an utterance from another language. An example of the Code-Mixed data can be seen in the Figure 1 \cite{r1}.
In countries where bilingualism is a common practice, we often see people naturally switching between the two languages. A significant challenge to research is that there are no formal sources like books or news articles in Code-Mixed languages, and studies have to rely on sources like Twitter or messaging platforms. Generating and evaluating the available or produced data without a baseline is primarily reliant on people who are fluent in both languages. 

Furthermore, present language models are ineffective in Code-Mixed situations, where morphemes, words, and phrases from one language are embedded in the other. As Code-Mixing has long been a way of communication in a multi-cultural, multi-lingual society, the next generation of AI bots should be able to understand Code-Mixed text.

The inherent challenges with the code-mixed data make the widely popular evaluation metrics like BLEU (Bilingual Evaluation Understudy Score) and WER (Word Error Rate) less effective. With the given task, the main objective is to propose and develop new strategies that address the overall need for quality evaluation of the generated Code-Mixed text.

This paper aims to assess the quality of the generated Hinglish text. The proposed model uses the transformer-based multi-lingual BERT \cite{r2} to obtain the embeddings of the Hindi, English, and Hinglish text. A similarity score is computed between synthetically generated Hindi and English text and human-generated Hindi and English sentences. This score will serve as the sentence's agreement or disagreement factor.

\begin{figure}[h!]
    \centering
    \includegraphics[scale=0.13]{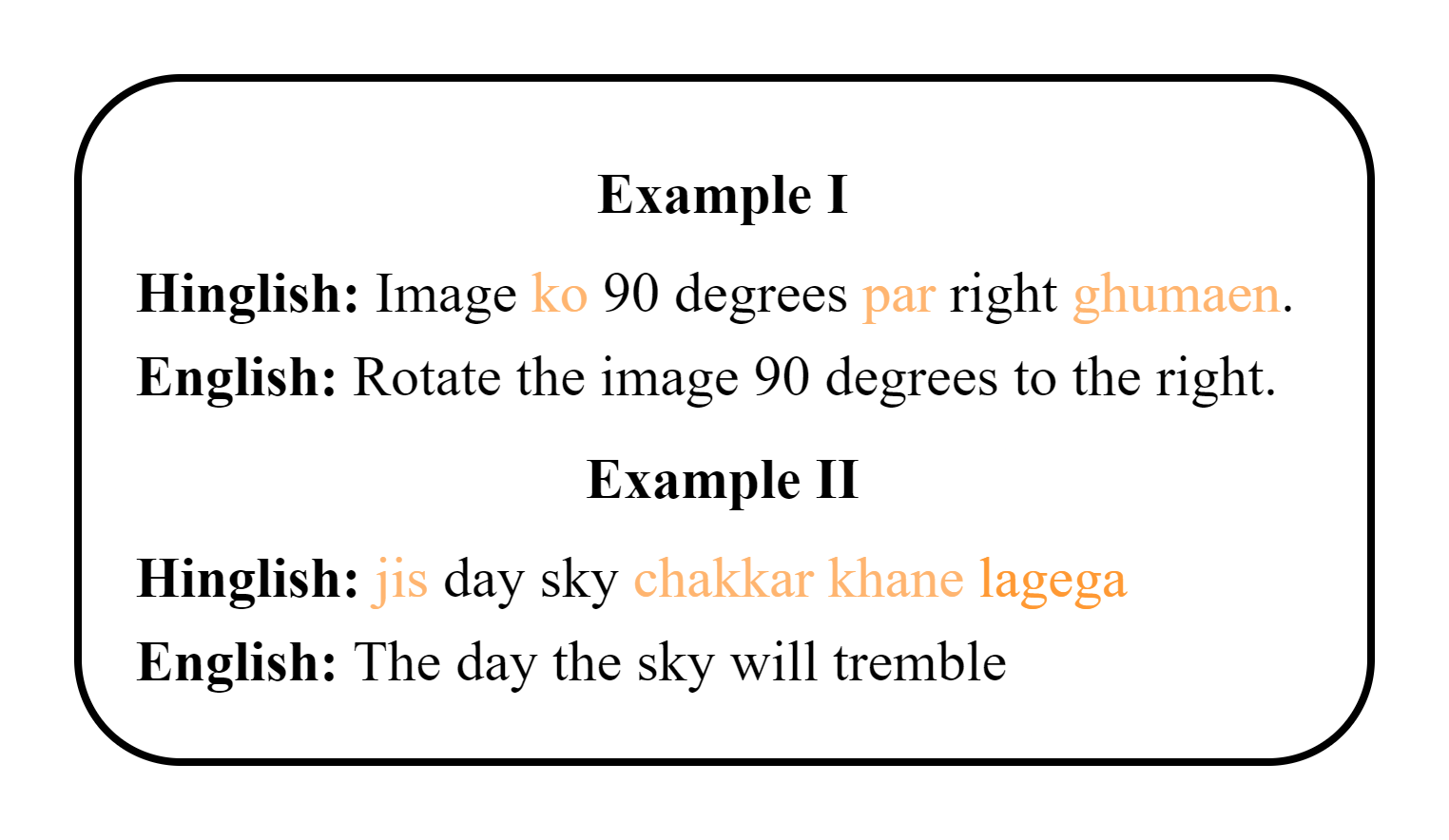}
    \caption{Examples of Hinglish and English sentences}
    \label{fig:my_label}
\end{figure}

\section{Related Work}
Developing methodologies and resources for diverse Natural Language Processing applications incorporating multi-lingual and Code-Mixed languages has recently become popular. Few of them are word-embedding \cite{r5}, question answering system \cite{r3,r13}, Code-Mixed text generation \cite{r6} and Code-Mixed language modeling \cite{r4}.

Various techniques were used to generate the Code-Mixed data, including matrix frame language theory, semi-supervised approach \cite{r9},  using dependency parsing \cite{r11}, equivalent constraint theory \cite{r6} and Generative Adversarial Networks \cite{r10}. A Metric Independent Evaluation Pipeline (MIPE) \cite{r14} considerably enhances the correlation among evaluation measures and human assessments of the generated Code-Mixed data. In the candidate's Hinglish phrase, MIPE minimises spelling differences and language switches for evaluation. Based on the significance of the terms missing from the candidate Hinglish sentence, deduct credit from the evaluation score. By arranging the candidates as well as the reference sentences into the phrases and using the paraphrasing ability, it also deals with the issue of having a restricted amount of reference sentences.

\section{Dataset}

HinGE\cite{r12}, a recently proposed dataset, is used for the HinglishEval task. The dataset contains 1976 English-Hindi sentence pairs and corresponding synthetic and human-generated Hinglish sentences.

\textbf{Human Generated Sentences:} Each English-Hindi sentence pair has at least two Hinglish sentences, with a total of 6694 such sentences.

\textbf{Synthetically Generated Sentences:}  Using two different algorithms(WAC and PAC), synthetic Hinglish sentences are generated for each English-Hindi sentence pair. Each sentence is then given a quality rating by two human annotators. There are 2766 such sentences. For each sentence, the rounded off average of the two quality ratings is provided under the label of  ‘Average Rating’ and the absolute difference of their scores is provided as ‘Disagreement’. An instance of the HinGE dataset \cite{r12} can be seen in Table 1.
\begin{table*}
\centering
\begin{tabular}{|c|c|c|c|c|}
\hline
\textbf{English} & \textbf{Hindi} & \textbf{{\textbf{\vtop{\hbox{\strut Human-Generated}\hbox{\strut Hinglish}}}}} & \textbf{WAC} & \textbf{PAC}\\
\hline
\multirow{3}{*}{\begin{tabular}[c]{@{}c@{}}The reward of \\ goodness shall\\ be nothing\\ but goodness\end{tabular}}  & \multirow{3}{*}{\begin{tabular}[c]{@{}c@{}} {\dn a+CAI kA bdlA}\\ {\dn a+CAI ke alAvA}\\{\dn aor kyA ho sktA h}?\end{tabular}} & \begin{tabular}[c]{@{}c@{}}The reward of\\achai shall be \\ nothing but achai.\end{tabular}         & \begin{tabular}[c]{@{}c@{}}reward ka \\badla reward\\ ke nothing\\ aur kya\\ ho sakta hai\end{tabular} & \begin{tabular}[c]{@{}c@{}}reward of\\ goodness\\ goodness\\ ke siva aur kya \\ho sakta hai\end{tabular} \\ \cline{3-3}
 &  & \begin{tabular}[c]{@{}c@{}}Goodness ka badla\\ goodness\\ ke siva aur kya ho\\ sakta hai.\end{tabular} & \multirow{2}{*}{\textbf{\begin{tabular}[c]{@{}c@{}}Rating1: 7\\ Rating2: 4\end{tabular}}}          & \multirow{2}{*}{\textbf{\begin{tabular}[c]{@{}c@{}}Rating1: 9\\ Rating2: 7\end{tabular}}}            \\ \cline{3-3}
 &  & \begin{tabular}[c]{@{}c@{}}Achai ka badla\\ shall be\\ nothing but achai.\end{tabular}  & & \\ \hline
\end{tabular}
\caption{
An Instance of the HinGE dataset
}
\end{table*}

\section{Proposed Model}
The proposed approach used a two-step procedure for both the rating and disagreement prediction tasks. The first step is to fine-tune multi-lingual BERT, a language model that has been pre-trained on 104 languages and is used to classify and evaluate disagreements further on. The second step used the deep semantic features obtained from multi-lingual BERT for various phrase categories to train a classifier neural network.
\begin{figure}
    \centering
    \includegraphics[scale=0.09]{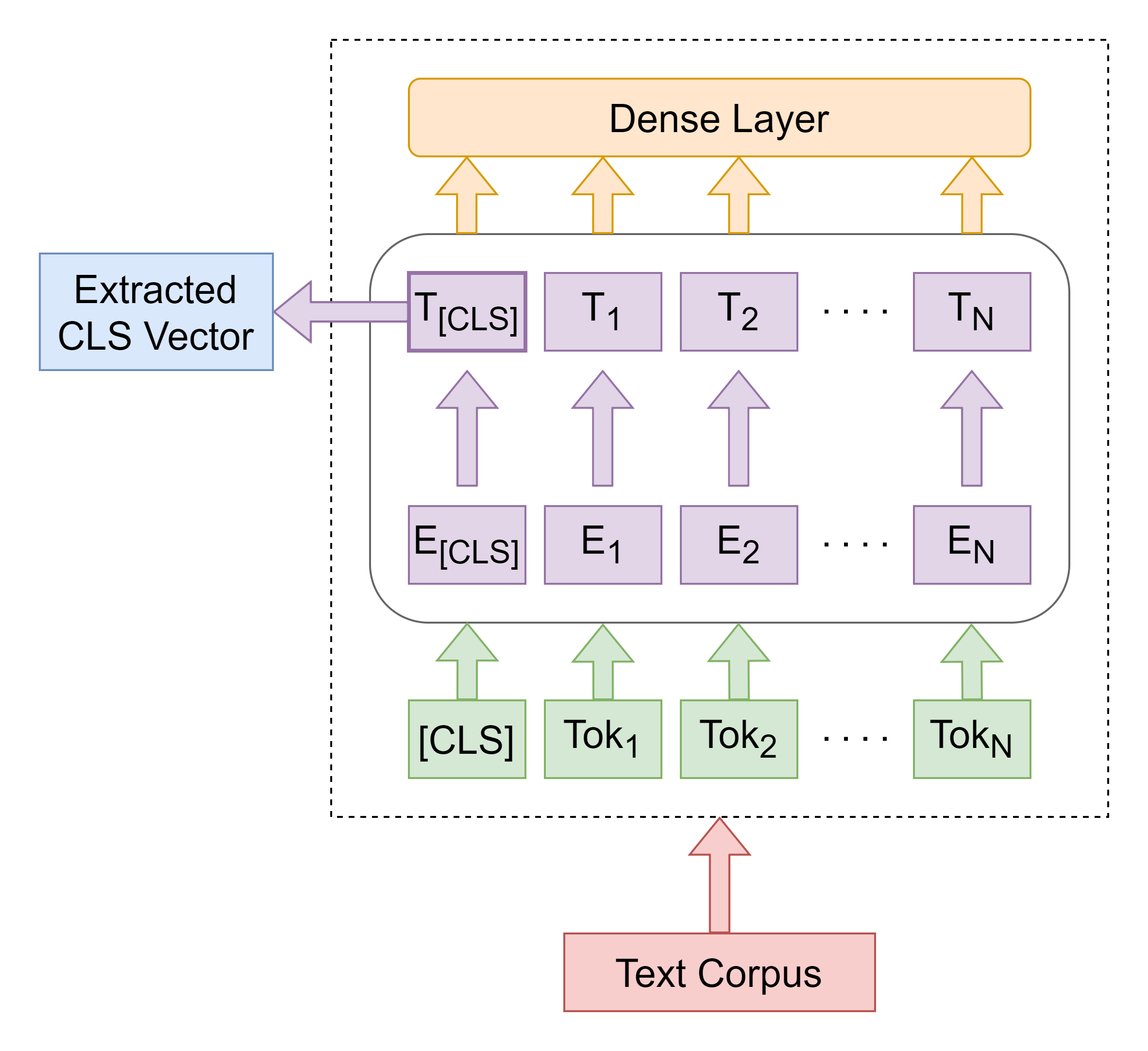}
    \caption{Extraction of CLS Vectors}
    \label{fig:my_label}
\end{figure}
\subsection{Multi-lingual BERT}
We employed the BERT-base-multi-lingual-cased model, a modified pre-trained BERT encoder that is pretrained in a self-supervised manner using the largest Wikipedias with the goal of masked language modeling.
Multi-lingual BERT allows us to provide it with two sentences as input in the form:

\medskip
[CLS] Sentence A [SEP] Sentence B [SEP]
\medskip

Here, the CLS and SEP tokens are special tokens that allow BERT to recognize the beginning of an input and a separation between two different input sections. The central SEP token ensures that BERT knows there are two different sentences in the input.

Based on the dataset, we had four models for different types of sentences—English, Hindi, Synthetic Hinglish, and Human Hinglish. We ran the language model four times for each sub-tasks with various sorts of sentence pairs each time. The first two models trained had Synthetic Hinglish sentences as Sentence B, with Sentence A being the corresponding English sentence for model 1 and Hindi sentence for model 2. The next two models did the same, with Human Hinglish sentences paired with English sentences for model 3 and Hindi sentences for model 4. This process is repeated twice for rating classification and again for disagreement classification. The proposed model employs the BERT AutoTokenizer to tokenize the inputs and the Adam optimizer to train at a learning rate of 1e-6 for five epochs.

We extracted the deep semantic text features from each model using the BERT source code. The CLS token's feature vector is extracted from the output results of the last hidden layer, which is a 768-dimensional deep semantic feature of the legal language. We chose the CLS token, also known as the Classification token, as it has the fixed embedding that appears at the beginning of every sentence. Since all words infer the output of this token in the phrase, this CLS vector provides BERT's understanding of the sentence, which is particularly beneficial for a sentence classification task. The extraction process can be seen in Figure 2, where T\textsubscript{[CLS]} is the CLS vector output of the last hidden layer of multi-lingual BERT extracted from the model.
\subsection{Classifier Neural Network} 
We obtained four sets of feature vectors from multi-lingual BERT for each sub-task. Two of them had dimensions of (2766, 768) and used synthetic Hinglish words as input. The other two used human-generated sentences and had dimensions of (6694, 768). To reduce the dimensionality of the latter, we averaged the vectors corresponding to each English-Hindi sentence pair resulting in a set of dimensions (1976, 768).

For each synthetic sentence, we concatenated the first two sets of vectors, and corresponding to each vector in this set, we appended its respective human vector. After combining the four-vector sets, we had an input of size (2766, 3072) for each sub-task.

The proposed model is trained with two fully connected neural networks, one for each sub-task, using these concatenated vector sets as their respective inputs. Both neural networks had two Linear hidden layers of dimensions (3072, 1536) and (1536, 768) with a final layer of size (768, 10). After each layer, a Rectified Linear activation function is being used. Binary Cross-Entropy loss is the chosen loss function. Adam optimizer were used along with a learning rate of 5e-6. Ten training epochs were used to train the disagreement classifier compared to just three for the ratings classifier. The entire procedure is illustrated in Figure 3, where the four concatenated CLS vectors are passed as input to a classifier neural network.
\begin{figure}[h]
    \centering
    \includegraphics[scale=0.10]{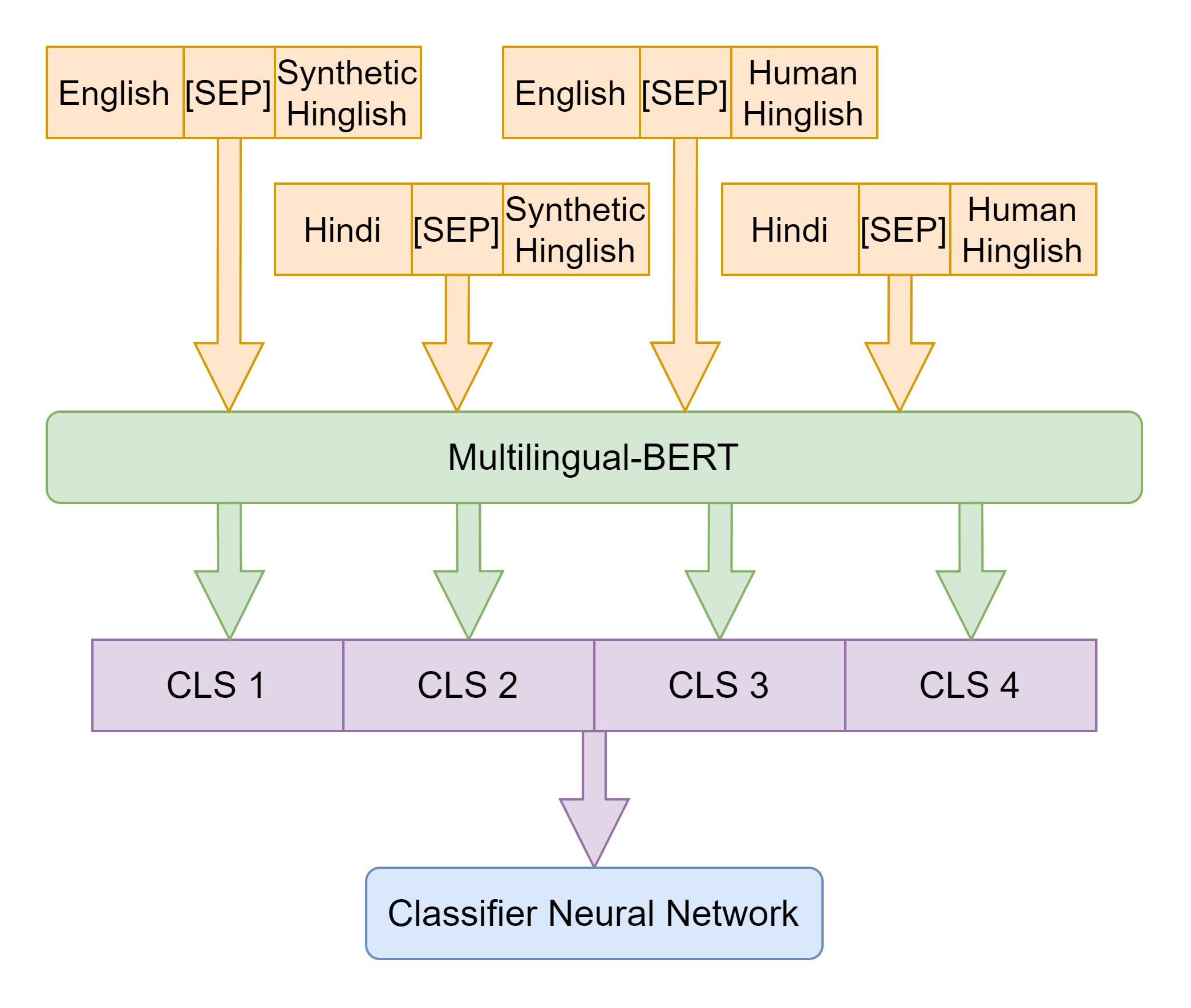}
    \caption{Proposed Model Architecture}
    \label{fig:my_label}
\end{figure}

\begin{table*}[t]
\centering
\begin{tabular}{llccccc}
\hline
\multicolumn{2}{l}{} & \multicolumn{3}{c}{Sub-Task 1(Average Rating)} & \multicolumn{2}{c}{Sub-Task 2(Disagreement)} \\ \hline
 & Model & F1 & CK & MSE & \ F1 & MSE \\ \hline
\multirow{2}{*}{Validation} & Baseline \footnotemark[1] & 0.09504 & -0.01530 & 15.000 & 0.15541 & 17.000 \\
 & \textbf{Proposed Model} & 0.23493 & 0.06515 & 3.000 & 0.19400 & 4.000 \\ \hline
\multirow{2}{*}{Test} & Baseline \footnotemark[1] & \multicolumn{1}{l}{0.26637} & \multicolumn{1}{l}{0.09922} & \multicolumn{1}{l}{2.000} & \multicolumn{1}{l}{0.14323} & 5.000 \\
 & \textbf{Proposed Model} & \multicolumn{1}{l}{0.21796} & \multicolumn{1}{l}{0.07337} & \multicolumn{1}{l}{3.000} & \multicolumn{1}{l}{0.24252} & 4.000 \\ \hline
\end{tabular}
\caption{Validation and Test results on the HinGE dataset}
\end{table*}

\section{Results and Evaluation}
\subsection{Results}
The evaluation metrics used are F1 Score, Cohen's Kappa (CK), and Mean Square Error (MSE). The submissions created by the model achieved rank 4 in rating prediction and rank 2 in disagreement prediction, not accounting for baseline scores, based on F1 Score, Cohen’s Kappa, and mean squared error. The proposed approach results can be seen in Table 2, compared with the baseline scores.

We attained F1 scores of 0.218 on Rating classification and 0.242 on Disagreement classification after training over 2766 synthetic sentences and testing over 791 synthetic sentences.

\footnotetext[1]{\url{https://codalab.lisn.upsaclay.fr/competitions/1688##results}}

\section{Conclusion and Future Work}

This paper used a two-step approach to solve the text classification problem. For each pair of phrase types, deep semantic text features were initially extracted using multilingual BERT as CLS vectors. These vectors were then properly combined, and  processed by a fully-connected classifier neural network. The results suggest the proposed model is useful and that the obtained results can be greatly improved by fine-tuning and training with larger data, which could be a future research direction.  

\bibliography{anthology,acl2020}
\bibliographystyle{acl_natbib}

\end{document}